\documentclass[10pt, a4paper]{article}

\usepackage[]{lrec-coling2024} % this is the new style
\usepackage{tipa}
\usepackage{adjustbox}
\usepackage{multirow}
\usepackage{xurl}

\title{Kallaama: A Transcribed Speech Dataset about Agriculture in the Three Most Widely Spoken Languages in Senegal}

\name{Elodie Gauthier, Aminata Ndiaye, Abdoulaye Guissé} 

\address{Orange Innovation, Jokalante SARL, École Polytechnique de Thiès \\
         Lannion, France, Dakar, Sénégal, Thiès, Sénégal \\
         elodie.gauthier@orange.com, amina.ndiaye@jokalante.com, aguisse@ept.sn\\
}

\abstract{
This work is part of the Kallaama project, whose objective is to produce and disseminate national languages corpora for speech technologies developments, in the field of agriculture. Except for Wolof, which benefits from some language data for natural language processing, national languages of Senegal are largely ignored by language technology providers. However, such technologies are keys to the protection, promotion and teaching of these languages. Kallaama focuses on the 3 main spoken languages by Senegalese people: Wolof, Pulaar and Sereer. These languages are widely spoken by the population, with around 10 million of native Senegalese speakers, not to mention those outside the country. However, they remain under-resourced in terms of machine-readable data that can be used for automatic processing and language technologies, all the more so in the agricultural sector.
We release a transcribed speech dataset containing 125 hours of recordings, about agriculture, in each of the above-mentioned languages. 
These resources are specifically designed for Automatic Speech Recognition purpose, including traditional approaches. 
To build such technologies, we provide %ready-to-use n-gram language models in Wolof and Pulaar along with the training texts, 
textual corpora in Wolof and Pulaar, and a pronunciation lexicon containing 49,132 entries from the Wolof dataset. 
 \\ \newline  }

\begin{document}

\maketitleabstract

\section{Introduction \label{sec:intro}}
%UNESCO report 
While information and communication technology is essential for many to thrive, 6 billion people still lack access to broadband, 4 billion lack access to the Internet, and 2 billion lack access to a mobile phone \cite{zelezny2018digital}. Latest estimations from \citet{uis2023} indicates around 213 million adults (population over 15 years old) who could not read or write, in 2022, across the sub-Saharan African region, including nearly 49 million young people (15-24 years old). \\
% In Senegal, 4\,226\,792 adults are concerned by illiteracy according to the same statistics. It nearly equals to a quarter of the population.
In Senegal, \citet{ansd2021poverty} reports an overall illiteracy rate of 48,2\%, reaching 62,7\% in rural area.

Literacy rate relates to the official language of a country. In Senegal, the official language is French % and is used in administration and education, 
but is seldom spoken by the population in their daily lives. Senegalese people primarily use their native languages or Wolof, as a vehicular language, to communicate. 
% Senegal : nearly 65 percent of those who do not use the internet are held back by deficiencies in digital literacy // près de 65 % des personnes qui n'utilisent pas l'internet sont freinées par des lacunes en matière de culture numérique
\citet{wdr2021data} reports that nearly 65\% of Senegalese who do not use the Internet are hindered by a lack of digital literacy. This is partly due to the limited (if not none at all) availability of content in the language they speak. Currently, there is a severe lack of accessible content for those who do not speak the official languages in Africa. 
The development of technologies and tools for the most widely spoken languages would enable a larger proportion of the Senegalese people to use smartphones and applications, and to access content that is still unavailable today. 

Research work as \citet{medhi2011appdesign} and the success of WhatsApp voice communication show that the development of conversational voice services in local languages is a credible and promising way of making services more accessible. \citet{aker2011dial} also suggested in that time that combining a voice-based approach with information that can be accessed through answers to common farmer questions would overcome literacy challenges due to the common texting modes. To make progress in this area, robust speech recognition systems need to be designed for these languages.
While automatic speech recognition (ASR) technologies tend to be mature in the languages most commonly found on the Web, there is still very few solutions dedicated to African languages.

In Senegal, Wolof, Sereer, Pulaar, Joola, Malinké and Soninké languages are recognised as national languages in the Constitution, but none of these six languages seriously benefit from the major technological advances generated by AI. Efforts have been made to develop speech resources and technologies in Wolof (see section \ref{sec:existing_res}) but no resources are available for Large Vocabulary Continuous Speech Recognition (LVCSR) in Pulaar nor Sereer. Yet, Wolof, Pulaar and Sereer languages are spoken in more than two-thirds of Senegal country \citep{leclercSenegal}.

% UNESCO 2018 report
Agriculture is the primary source of income for 2 billion people around the world \citep{zelezny2018digital}. In Senegal, 55\% of the population is involved in the agricultural value chain, including family farming, livestock breeding, and fishing. 
Today, digital technologies are assisting farmers in expanding their businesses by enabling them to position themselves on marketplaces, providing them with information on commodity prices, and granting them access to suitable financial services. Nonetheless, such solutions are still not appropriate for farmers, particularly given the prevalence of written communication and the use of a language they do not speak, when interacting with these interfaces.\\
With the intention of speech solutions development, this dataset is intended to fill the gap in this area.

\paragraph{Paper contribution.} 
% This paper presents the curated data sets, both concerning speech and texts, to build a machine-readable dataset that can be used to develop voice-based solutions in Wolof, Pulaar and Sereer languages. To the best of our knowledge, this transcribed dataset is the largest ever created for these languages. Moreover, Pulaar and Sereer languages were, until now, never address in large-scale speech processing. 
This paper presents the dataset created from speech data produced and annotated during the Kallaama project, as well as textual data gathered from the web, with the aim of developing voice-based solutions in local languages.
\paragraph{Paper outline.} After an introduction, we present the project in section \ref{sec:Kallaama}. The targeted languages are described in section \ref{sec:target_lang} and existing resources in these languages are listed in section \ref{sec:existing_res}. Section \ref{sec:methodo} presents the collection methodology, while section \ref{sec:dataset} gives details about the dataset. Then, we present some of the challenging times we faced during the project in section \ref{sec:challenges} and we mention some of the resulted limitations in section \ref{sec:limitations}. Finally, section \ref{sec:conclu} concludes and gives some perspectives about the use of Kallaama.

\section{The Kallaama project \label{sec:Kallaama}}
"Kallaama" means "speech" (from Latin "verbum") in Wolof.
\subsection{Description}
%Le projet Kallaama vise à combler ce manque et à produire 60 heures de données audios transcrites et annotées pour entraîner des systèmes de reconnaissance de parole dans 3 des principales langues nationales du Sénégal : wolof, sérère et pulaar, soit 180 heures de parole localisée.
As mentioned in section \ref{sec:intro}, no resources are available to build LVCSR systems in Pulaar nor Sereer. Only a small amount exist in Wolof, but none focus on agriculture.
The Kallaama project aims to fill this gap by producing several dozen hours of transcribed and annotated localized audio data, to train speech recognition systems in three of the Senegal's main national languages: Wolof, Sereer and Pulaar.

%Le choix de ces 3 langues est guidé par le nombre de locuteurs concernés dans le pays. Wikipédia mentionne 16 millon de locuteurs wolofones dont 8 M de locuteurs natifs ainsi que 3,5 millon de locuteurs pulaar et 1,2 million de locuteurs sérère. Ce sont, de fait, les 3 langues vernaculaires les plus représentées au Sénégal. En choisissant le pulaar et le sérère en sus du wolof, la proportion de population sénégalaise que l’on peut imaginer atteindre sera largement démultipliée. D’autant plus que ces langues traversent quelques frontières.
The choice of these 3 languages was guided by the number of speakers in the country. There are around 5 million native speakers of Wolof, 3.5 million native speakers of Pulaar and 1.3 million native speakers of Sereer \citep{leclercSenegal}, which represent three quarter of the total population. These are the 3 most widely spoken languages in Senegal, and they are also spoken cross several borders.

%Les données produites par Kallaama sont des énoncés naturels et spontanés, avec du vocabulaire en contexte, destinées à développer des modèles de reconnaissance de parole à grand vocabulaire, en particulier relatives au domaine agricole. La reconnaissance de parole est le principal verrou technologique à lever pour développer des services vocaux (voicebots, callbots…) au bénéfice des personnes peu ou pas lettrées.
The data produced are natural, spontaneous utterances, with vocabulary in context, designed to develop large vocabulary speech recognition models, particularly relating to the agricultural domain. Speech recognition is the main technological barrier to be overcome to develop voice-based services for people with little or no literacy. Agriculture plays an important role in rural activities in Senegal. It is one of the pillars of the Senegalese economy, estimated to contribute 15\% of GDP in 2022 as mentioned in the Annual Agricultural Survey of \citet{agri_report_2023}, and a large proportion of the population remains directly dependent on it.

\subsection{Use case}
Several and local companies and start-ups in the IT sector are increasingly embarking on the production of AI solutions that take national languages into account. These are essentially automatic text or speech translation solutions, allowing them to expand their customer base and offer their services in French or English to users who prefer Wolof, Pulaar, Sereer, or other languages. 
Serious initiatives also have been noted in the development of multilingual chatbots and voicebots. However, due to the scarcity of natural language data in local languages, most of them rely on synthetic data from machine translation systems. %the big challenge is to produce these resources as we do in the Kallaama project, but also and above all to develop language models like BERT and GPT-like initiatives, focused on the professions (agriculture, health, telecommunications, marketing, etc.) of the targeted AI solutions.
% AG contribution: Also, the datasets produced allow the scientific community to popularize local languages and to focus on the production of language models specific to business digital needs. This makes it possible to train machine and deep learning models in order to facilitate the exploitation of masses of high value-added resources, produced (texts, images, audio and videos) in real time in call centers, in social networks, in discussion forums, etc. NLP and AI solutions are still necessary for machine translation and for implementing knowledge bases, conversational assistants, recommendation systems or decision-making support.
Moreover, AI models still only marginally address agriculture. Yet, digital solutions for agricultural extension work cover a range of needs, including information delivery services, small business management tools, training and skills enhancement, and financial services \cite{zelezny2018digital}. \\

The Kallaama dataset contributes to the growth of the agricultural sector in Senegal. It can strengthens food security by providing vital information directly in the farmers' native language, through the development of voice-based services such as personalised agricultural and financial advice to smallholder farmers.
Besides, the produced transcriptions increase the available datasets in Senegalese languages, and will boost AI-based developments for agriculture, like setting up knowledge bases, conversational assistants, recommendation systems and decision support systems. 

\section{Focus on the targeted languages \label{sec:target_lang}}
Wolof, Pulaar and Sereer languages are spoken by nearly 80\% of native speakers in Senegal. \citet{cisse2005langues} indicates 43,7\% of Wolof native speakers, 23,2\% of Pulaar native speakers and 14,8\% of Sereer native speakers. These three languages belong to the Niger-Congo phylum and are part of the North-Atlantic family group. They are toneless, unlike most Niger-Congo languages\footnote{As mentioned by \citet{creissels2019morphology}, non tonal languages are primarily spoken in the Atlantic languages of western Senegal and the Bantu languages of eastern Kenya and Tanzania (like Swahili).}. 
By having a national status, the three languages received an official spelling system. It is based on the Latin characters.\\

\begin{figure}[t]
    \centering
    \includegraphics[width=.9\linewidth]{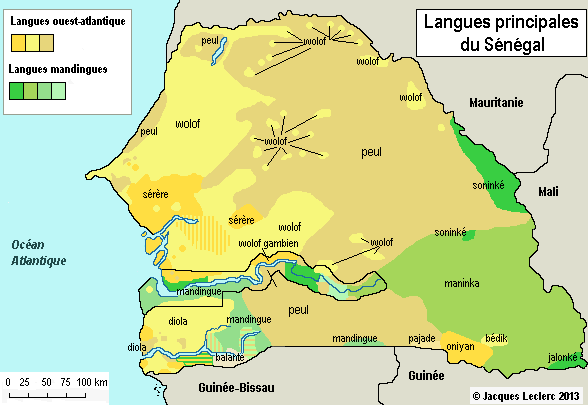}
    \caption{Map of main languages spoken in Senegal \citep{leclercSenegal}}
    \label{fig:lang_SN}
\end{figure}

\subsection{Wolof\label{subsec:wolof_description}}
Wolof is by far the most spoken language in Senegal. It is the native language of about 5 millon speakers. The Wolof spoken in Senegal is identified by the ISO 639-3 language code name "\texttt{wol}"\footnote{Another variant of Wolof is spoken in Gambia, for which the ISO code is "\texttt{wof}".}. 
%Even if French have the official language status, informations in the public space are more and more delivered in Wolof (\citealp{\citealp{}; }). Radio and TV shows broadcasted in Wolof are steadily increasing.
%All across Africa, national languages are becoming more prevalent in comparison to the official languages \citep{voisin2017afrique} and, in Senegal, the use of Wolof is a good illustration of this. 
% Despite French is the official language of the country and is used in administration, education, and the media, Wolof is the predominant language used in practice. 
By being spoken by almost 90\% of the population, Wolof is the national language of communication, widely surpassing French in terms of usage \citep{cisse2005langues}. On social networks, comments are mainly written in Wolof in response to articles written in French. The National Assembly provides a translation service as 20\% of MPs do not speak French. Additionally, private TV and radio channels have developed programmes in Wolof \citep{oif2022usage}.
% Several African languages are now becoming more prevalent in comparison to the official languages \citep{mous2003loss}

% The alphabet is composed of 25 consonants and 16 vowels. 
% Wolof is characterized by a length contrast within the vowels.
% It terms of syntax, Wolof follows a SVO word order.

\subsection{Pulaar\label{subsec:pulaar_description}}
% https://www.jeuneafrique.com/138138/societe/plan-te-peule-rencontre-avec-un-peuple-sans-fronti-res/
% https://lgidf.cnrs.fr/sites/lgidf.cnrs.fr/files/images/PEUL_24_06_19_A4.pdf
Pulaar is part of the Fulfulde languages. Fulfulde is spoken in about 20 sub-Saharan African countries, by nearly 30 million people, from West to Central Africa.
"Pulaar" refers to the variant spoken in Senegal. Pulaar is the native language of about 3.5 millon of the Senegalese people, making it the second most widely spoken language in the country. % Northern and Southern Pulaar speakers do not always understand each other.
Pulaar speakers across the country do not always understand each other. Clear differences in accents and lexicons should be noted. There may be borrowings and mutual influences between the accents. 
The Pulaar spoken in the north, considered as the reference in Senegal, used in the areas of Fouta Toro and Ferlo, is different from the one spoken in the south, in Fouta Djallon and Boundou areas, and from the centre (particularly in Saloum).
The economic activities practiced by the Fulani in these regions are at the origin of these differences, without forgetting the mobility of populations and inter-cultural exchanges.

\subsection{Sereer\label{subsec:sereer_description}}
Sereer language is spoken by around 1 million speakers, making it the third language spoken in Senegal. 
Several dialects are spoken in Senegal \cite{renaudier2012derivation}, and mutual understanding between Sereer speakers is sometimes difficult. The majority of the recordings proposed in this dataset are in Sereer-Siin (ISO 639-3 code "\texttt{srr}") variant, which is spoken in a region between the Petite Côte (south of Dakar) and the Gambia, and which is considered as "standard" Sereer. Nonetheless, depending on where the recording was made, it may be in another variant.
The official script is based on the standard Sereer-Siin variant but is very little used for writing. The language is fundamentally spoken.
% \paragraph{Phonology.} The phonemic system of Sereer is composed of 5 short vowels with long counterpart, and is composed of 20 consons.

\section{Existing language resources for Wolof, Pulaar and Sereer \label{sec:existing_res}}
More material (in any field of application, from linguistic description to language learning) can be found in Wolof, as a vehicular language. The situation is very different with Pulaar and Sereer: as vernacular languages, they are mainly spoken and rarely written. The presence of Wolof online is preponderant against Pulaar and Sereer, reflecting its place in the Senegalese society.

To build voice-based solutions, very few datasets were released so far in Wolof. Pulaar and Sereer speech datasets are nearly non-existent, exceptions made from the initiatives presented below. \\
Before this work, the largest transcribed speech dataset in Wolof was the one collected by \citetlanguageresource{GAUTHIER16.460}. It consists in 18 hours of validated read sentences. \\
Wolof is also proposed in FLEURS, a multilingual dataset consisting in translation of English sentences that has been read by native speakers \citeplanguageresource{conneau2023fleurs}. \\
\citetlanguageresource{waxalMultilingual2022} conducted a project of large collection of Wolof speech consisting in 519 hours of audio recordings, for which 6.45 hours have been transcribed so far. Among the 2,018 Wolof transcriptions, we counted 608 translations in Pulaar, 571 in Sereer but only 156 audio recordings translated in both languages. \\
Finally, the last significant work we found involving the three Senegalese languages addressed in this paper, is part of a data collection project of isolated words for keyword spotting, led by the Senegalese Galsen AI community \citeplanguageresource{waxalGalsenAI2022}. %The with African Languages project does exist, but its datasets are limited to the recognition of isolated words (\url{https://www.k4all.org/project/keyword-spotting-with-african-languages/)}

\section{Collection methodology \label{sec:methodo}}
\subsection{Audio recordings and transcriptions\label{subsec:speech_set}}
\textbf{Audio recordings.} The recordings are about agriculture. The recorded consist of farmers, agricultural advisers, and agri-food business managers. All the data is produced by Jokalante, a Senegalese company specialising in the dissemination of information about agriculture in local languages. Type of recordings comprise interactive radio programmes, focus groups, voice messages, push messages and interviews\footnote{For each dataset, the number of recordings per programme type is detailed in appendix \ref{app:program_types}.}.
Therefore, spontaneous speech is prevailing. Quality of audio may vary depending on the type of programme. For instance, focus group are made outdoor and so noises may arise from the outside (cars, wind, birds, additional voices). In radio programmes, music and jingles sometimes also appear. 
A selection from these recordings were transcribed, resulting in over a hundred hours of spontaneous speech in the three targeted languages (see Section \ref{subsec:speech_dataset_details}). \\
\textbf{Transcriptions.} To produce written form of the audio recordings, we asked the transcribers to follow the rules edited by the Centre of Applied Linguistics of Dakar (CLAD)\footnote{\url{http://clad.ucad.sn/}}, which coordinates the orthographic standardization of the national languages in Senegal. Despite that caution, it was very difficult to obtain a standard form in the writing of languages concerned in the present work. As mentioned by \citet{robert2022wolof} for Wolof, official rules are rarely used by the population (as example, advertisements are often written with alternative forms), even if an official orthography is established since 1971. The same situation appears for Pulaar and Sereer, and this is primarily due to the fact that the national languages are taught very little in the education system. In addition, the transcription work involved recordings of spontaneous speech, making the work all the more complex and time-consuming. %Spontaneous speech is informal and unrehearsed speech produced by a speaker in a dynamic and casual manner.
Transcription task was performed by 3 students in Linguistics, in the language they natively speak. They used the dedicated Transcriber\footnote{\url{http://trans.sourceforge.net/}} tool to achieve the task. The work took 9 months to complete.
Then, 3 qualified experts, specialised in the languages of the transcripts, reviewed a subpart of the transcriptions produced by the students. At first, we were aiming to verify half of the transcriptions produced, for each language. 
But it was an ambitious goal given the complexity and arduousness of the task required. 
Nonetheless, nearly 13 hours of speech transcription were checked in Wolof, 11 hours in Pulaar and 11 hours in Sereer within the allotted time.

\subsection{Texts collection\label{subsec:texts}}
Senegalese languages are low-resourced. Very few data in the targeted languages were unearth.   
First of all, no documents on agriculture were found. \\
% Unsurpringly, the more data we found was in Wolof language. As being a vehicular language, most people (whether they are native or non-native speakers) use it for written communication. Pulaar and Sereer, in contrast, are mostly spoken and so is quite nonexistent under a written form on the Web. 
Since the observations from \citet{renaudier2012derivation} who mentioned that the local press is predominantly written in French, with only a few newspapers available in Wolof, and that no press were available in Sereer at the time, the situation remains unchanged.

% Web scraping methods were used to collect the textual data on the Web. We also found open source dataset
Most of the written sources found were in Wolof. 
The Wolof corpus we distribute is composed of the books, Wikipedia articles (dump from summer 2023), the first book of the New Testament, two historic blogs about Hubert Fichte (a German novelist) and Cheikh Ahmadou Bamba (a theologian), publicly available online articles from newspapers. Open source data from the Programme Algorithme et Solution (PAS) challenge\footnote{\url{https://www.ias.sn/pas}}, organised by the Institution des Algorithmes du Sénégal (IAS), have also been included. 
%and we also exploited 
Written data in Wolof can also be retrieve from open source research projects (in particular, the ALFFA project\footnote{ \url{https://github.com/getalp/ALFFA_PUBLIC}} and Masakhane\footnote{\url{https://github.com/masakhane-io/masakhane-ner/}}% https://github.com/masakhane-io/masakhane-ner/tree/main/text_by_language/wolof : 910 sentences and 44812 words from http://defuwaxu.com/ and http://saabal.com/
). We choose not to add them to our release as they are already clean and easy to get.

We found a very little amount of writings in Pulaar\footnote{Without distinction of dialectal variants spoken in Senegal.}. 
We extended our data research to include varieties spoken in regions bordering Senegal, and finally found more websites written in this language, particularly in Mauritania. 

About Sereer, although it is the third most widely spoken language in Senegal, gathering written data poses a significant challenge. Despite extensive research, no textual content was found on the consulted websites. We even went to the two main university of linguistic and language libraries in Dakar (Cheikh Anta Diop University (UCAD) and to the Institut Fondamental d'Afrique Noire (IFAN)), and only found two books written in Sereer. We still tried to apply some Optical Characters Recognition (OCR) tools to convert it into digital texts, but the special characters existing in Sereer were not recognised. \\

We have deliberately excluded all social networks in order to avoid biases that could be induced in future models. For example, \citet{dione2016contacts} observed, in her study on the online usage of Wolof and Sereer languages, that most internet users use French and Wolof alternatively in a single message. Besides, Wolof and Pulaar are the only two national languages present on the websites consulted by the author, while Sereer is also his subject of study. Moreover, the author indicates that internet users use Wolof to criticise, to display political choices and connivance, and to insult. For all these reasons, we preferred not to collect textual data from the forums. \\

These text corpora can be used to train monolingual and multilingual language models on the theme of agriculture. 
Language models are involved in various NLP tasks, such as ASR rescoring or natural language understanding/generation (NLU/NLG) modelling.

\subsection{Lexicon}
We found no dictionary with word pronunciation for Pulaar and Sereer, so we could not train a grapheme-to-phoneme (G2P) model for these languages.
For Wolof, we used the lexicon from ALFFA project to train a G2P model, in order to generate phonetic transcription of the Wolof speech set. The G2P model was trained using Phonetisaurus\footnote{\url{https://github.com/AdolfVonKleist/Phonetisaurus}}. The generated phonetic symbols are in X-SAMPA alphabet. 
We provide the G2P model and the lexicon in a GitHub repository. It can be useful to train HMM-based ASR models.

\section{Dataset details \label{sec:dataset}}
The dataset is released under the CC-BY~4.0~license. 
All textual data (transcriptions, text corpus, lexicon) are available on GitHub\footnote{\url{https://github.com/gauthelo/kallaama-speech-dataset}}. Audio recordings will soon be hosted on both OpenSLR and Zenodo platforms. Once released, URLs will be given in the GitHub repository.
\subsection{Audio recordings and transcriptions\label{subsec:speech_dataset_details}}
Audio files have been converted into 16\,kHz, 16-bit, mono channel, to fit the standard format used in ASR. 
Transcriptions are provided under the original Transcriber format (.trs), as well as in stm NIST format (.stm) as this one is more often used by ASR toolkits.\\
Details about speech datasets are given in table \ref{tab:speech_corpus_tab}.
\begin{table}[!h]
    \centering
    \begin{adjustbox}{width=1\columnwidth}
    \begin{tabular}{l|r|r|r|r}
        \multirow{2}{*}{\textbf{Language set}} & \multirow{2}{*}{\textbf{Total Duration}} & \multirow{2}{*}{\textbf{\#Turn-taking}} & \multicolumn{2}{c}{\bfseries Gender (\%)} \\
    & & & \textbf{F} & \textbf{M}  \\
        \hline
        Wolof & 55h12 & 46,907 & 10.2 & 89.8 \\
        Pulaar & 31h55 & 16,558 & 13.6 & 86.4 \\
        Sereer & 38h12 & 9,007 & 28.0 & 72.0 \\ \hline
        \textbf{Overall} & 125h19 & 72,472 & 17.3 & 82.7 \\
    \end{tabular}
    \end{adjustbox}
    \caption{Kallaama speech corpus overview}
    \label{tab:speech_corpus_tab}
\end{table}

%TODO: Est-ce qu'on peut dire quelque chose sur la différence de tours de parole entre les langues ? Notamment Pulaar vs Sérère qui présente un volume assez équivalent ? Est-ce que ça vient des émissions qui sont différentes ?
The high number of turn-taking that can be observed in the table \ref{tab:speech_corpus_tab} for the Wolof set is explained by a larger amount of interviews and focus group, involving more people in the talk. \\
The underrepresentation of women's voices in this corpus is regrettable, but it reflects the interviews conducted and the women's presence in agribusiness. \\
More details are given in Appendix \ref{app:speech_datasets_details}, where we also describe the checked subpart of the dataset.

\subsection{Texts collection}
The set of texts collected in Wolof, before the application of post-processing methods, totalled 3,244,642 words. 
The set of texts collected in Pulaar, before the application of post-processing methods, totalled 5,462,823 words.
As we said in subsection \ref{subsec:texts}, no written data were found in Sereer. \\
During post-processing non roman characters were removed from raw texts. Punctuation has been preserved to give users greater freedom, depending on how the corpus will be used. Finally, a new line was added after each final punctuation mark (the dot, exclamation and question marks) while spaces was added between other kind of typography mark (such as comma, colon, semi-colon, dash, bracket, etc.).  \\
After these post-processing steps, the Wolof text corpus contains 1,140,508 words, while the Pulaar text corpus contains 742,024 words. Detailed are given in table \ref{tab:wol_text_details} and table \ref{tab:fuc_text_details}, for Wolof and Pulaar respectively. 
This considerable reduction in content reflects the significant presence of other languages in the writings, particularly French and Arabic (we did not apply a language identification algorithm, but we did remove many characters in the Arabic alphabet). We also found a quite large number of Cyrillic characters in the collected texts from Wikipedia. \\
The compiled data will enhance the understanding of the usage of the languages and strengthen the ability to develop more robust linguistic tools. It will also serve as a training baseline for language models.

\begin{table}[!h]
    \centering
    \begin{tabular}{l|r|r}
        \textbf{Sources} & \textbf{\#Words} & \textbf{Distribution} \\ \hline
        Newspapers & 571,122 & 50\% \\
        Wikipedia & 346,604 & 30\%\\
        PAS Challenge & 157,119 & 14\% \\ 
        Book & 27,283 & 2\% \\ 
        New Testament & 22,468 & 2\% \\
        Blog & 15,912 & 1\% \\ \hline
        \textbf{Overall}  & \textbf{1,140,508} & \textbf{100\%} \\
    \end{tabular}
    \caption{Details about the web scrapped texts in Wolof, after cleaning}
    \label{tab:wol_text_details}
\end{table}

\begin{table}[!h]
    \centering
    \begin{tabular}{l|r|r}
        \textbf{Sources} & \textbf{\#Words} & \textbf{Distribution} \\ \hline
        Newspapers & 698,400 & 94\% \\
        Blog & 43,624 & 6\% \\ \hline
        \textbf{Overall}  & \textbf{742,024} & \textbf{100\%} \\
    \end{tabular}
    \caption{Details about the web scrapped texts in Pulaar, after cleaning}
    \label{tab:fuc_text_details}
\end{table}

\subsection{Lexicon}
In the aim to build ASR systems, we also provide a pronunciation dictionary for Wolof. 
It contains 49,132 phonetised entries from speech transcriptions and texts. Entries can also be loanwords, such as French words, since code-switching is frequent in Senegal and therefore occurs in the speech dataset.
Entries are phonetically transcribed with the X-SAMPA characters.

\section{Challenges encountered \label{sec:challenges}}
% La réalisation des transcriptions a été significativement ralentie en raison de l'absence de certains caractères spéciaux sur les claviers standards. Parmi ces caractères, on retrouve des lettres spécifiques telles que le Ɓ (B majuscule crocheté), ƴ, ɗ, ɓ, qui sont fréquemment utilisées dans la transcription des langues spécifiques, notamment le pulaar et le sérère. Face à cette lacune, les linguistes impliqués dans le processus de transcription se sont retrouvés dans l'obligation de copier ces lettres à partir de sources externes, notamment Wikipedia, ce qui s'est avéré être une tâche fastidieuse.
% Une solution a été trouvée en téléchargeant le clavier visuel SenLangEdit. Ce clavier spécialisé permet d'accéder plus facilement aux lettres nécessaires, simplifiant ainsi le processus de transcription et contribuant à accélérer le travail des linguistes.
Transcribers struggled to write some of the words because of the absence of certain characters on standard keyboards, such as \texthtb{} (b-hook),  \texthtc{} (c-hook), \texthtd{} (d-hook), \texthtp{} (p-hook),  \texthtt{} (t-hook) which exist in the spelling of African languages, in particular in Pulaar and Sereer. The SenLangEdit visual keyboard application\footnote{\url{https://esp.sn/senlangedit-un-clavier-virtuel-pour-la-promotion-des-langues-nationales/}}, especially developed to write the national languages of Senegal, still eased the transcription process.

\subsection{Writing rules}
It was particularly hard to find qualified experts for checking the quality of the produced transcriptions.
We ask each expert to make a report of their reviews. 
For Wolof, the expert declared that the work was quite easy. To complete the work within the allocated time, he managed to check nearly 13 hours of speech transcriptions and the conclusion was very encouraging. He noted a very good quality of work, with very few mistakes. In contrast, the two qualified experts hired to review the transcriptions in Pulaar and Sereer declared a tedious work. In spite of this, they manage to verify around 11 hours of audio recordings each. They raised numerous mistakes and warned us that their work would be more about rewriting than simple checking and correction. We detail the main mistakes found in Appendix \ref{app:mistakes_details}. \\  \newline 
% A lesson that can be drawn from this work, and which is worth noting, is that 
In fact, this assessment of the transcriptions quality primarily indicates a lack of written skills rather than a lack of attention to transcription quality. This is the result of attempting to transcribe a language that has traditionally been unwritten. As long as these languages are not taught to be write, there will be no good written productions.

% Moreover, a lesson that can be learned is that, during interviews, one should be care about the writing skills and not take speakers at their word. Native speakers say they know how to write their language, hearing perception can be a factor in errors for an untrained person. Plus, being able to speak a language does not necessarily mean knowing how to write it according to the established rules.

\subsection{Spoken dialects}
% Geographic areas
% Diouroup, Fatick, Kaffrine, Kaolack, Kebemer, Kelle, Koungheul, Louga, Matam, Ndoulo, Niodior, Nioro, Podor, Saint-Louis, Tambacound

The fact that recordings are produced throughout the Senegal\footnote{Diouroup, Fatick, Kaffrine, Kaolack, Kebemer, Kelle, Koungheul, Louga, Matam, Ndoundour, Niodior, Nioro, Podor, Saint-Louis, Tambacound.} made the transcription work much more complex, because of the several dialectal variations of Pulaar and Sereer which are spoken in the country (see sections \ref{subsec:pulaar_description} and \ref{subsec:sereer_description}). %For Pulaar, dialectal variations include Fulakunda in southern Senegal and Pulo Dieri in the north, while Sereer presents distinctions such as Sereer Sine, Sereer Mont Roland and Sereer Safeen. 
We selected the audio recordings at the very beginning of the project, before the transcriber hiring. But 
Pulaar or Sereer transcribers sometimes listened to programmes recorded that they did not understand because of a conversation in a dialect that they do not speak. Consequently, %they were be faced to misunderstanding and so mistranscription. 
we had to carry out a second recordings collection campaign that took into account the specific dialects spoken by the transcribers. \\
% This process highlights the need to take account of the semantic subtleties induced by different dialects, by illustrating the challenges inherent in accurately and exhaustively preserving the meaning of words in these languages.
This process also highlights the need to take account of the semantic subtleties between dialectal varieties, especially when dealing with a particular subject (in this case agriculture, but it could be health or finance) and illustrates the challenges inherent in accurately and exhaustively preserving the meaning of words in these languages.
% Amina contrib : the diversity of dialects in Pular and Sereer is a real difficulty, leading to a selection of content in Pular spoken in the north of Senegal to the detriment of Pular spoken in the south of the country. The result of this process is a reduction in the number of hours required for transcription. The same phenomenon was observed for Sereer, with dialects in central and eastern Senegal. Sereer Sine and Pular Djéri were used as part of the kallama project. 

\section{Limitations \label{sec:limitations}}
Transcription work is a very challenging task, and to produce a transcript from spontaneous speech, when overlapping events occur, sometimes in noisy environments, is even more so. 
Add to this the use of specialised software that is unfamiliar to the workers, with keyboards that are not adapted to writing the language, and the start of the work becomes even more tedious. \\
Despite the care of all the workers involved in this project, this dataset contains some transcription mistakes, and the spelling used may not correspond exactly to the expected standards, as pointed out in Section \ref{sec:challenges}. %The time spent by the linguistic experts during the qualification task, at the end of the work, showed that some of the transcriptions were not error-free. 
Only one transcriber was selected per language to carry out the transcription work. Perhaps some mistakes could have been avoided if more transcribers were doing the job (supposing this is possible, since the number of skilled people is very limited). But, due to production costs, we have chosen to provide the community with a larger set of transcribed data rather than increasing the number of transcribers. 
In this way, we have been able to increase the number of subjects covered on agriculture. A larger speech dataset is also more suitable for large-scale studies, such as phonetic and phonological research on Atlantic languages, a field where works lack.

\section{Conclusion and Opportunities \label{sec:conclu}}
% Conclusion
In this paper, we present the work carried out to create a transcribed speech dataset on Wolof, Pulaar and Sereer, the 3 most widely spoken languages in Senegal. This dataset comprises 55h of audio recordings in Wolof, 32h in Pulaar and 38h in Sereer, all along with their corresponding transcriptions. We also provide more generic text corpora in Wolof and Pulaar, as well as a Wolof phonetic lexicon along with its G2P model. These resources can be used for setting up traditional ASR systems. \\
As pointed out by many recent studies \cite{joshi2020state, van2022writing, ruder2022bias, adebara2022afrocentricNlp}, a lot of languages with large speaker populations still are under-represented in natural language processing (NLP) studies and applications, reinforcing inequalities such as knowledge access. 
We hope that this work will stimulate interest in the development of applications that incorporate the vernacular languages of Senegal, but also that it will be a source of inspiration and encouragement to develop the same kind of resources in order to progress towards the inclusion of languages in the world of AI. \\ \newline 
% Opportunities
Opportunities offered by this dataset are numerous.
%% Scientific: analyses phonétiques, analyses en situation conversationnelles (tours de parole, disfluences, etc.), analyses de l'impact de la parole spontanée et bruitée sur les systèmes de reconnaissance de parole, etc.
From a scientific perspective, the speech dataset released can be exploited for instance, to study phonetic phenomena occurring in a spontaneous context, to study speech interaction, or to study the impact of spontaneous and noisy speech on recognition systems.
%% Technique:
From a technical perspective, this dataset can be used to solve various AI tasks, including speech modelling (such as speech-to-text or spoken language understanding), automatic response modelling (as QA answering), and language modelling (used from scratch or used to fine-tuned a pre-trained multilingual model).
%% Techno:
From a technological perspective, it can be utilised to develop speech recognition systems, generic or specific to the agricultural sector, as well as localised conversational agents for answering questions on agricultural topics related to the Senegal context and in national languages.

\section{Acknowledgements \label{sec:acknowledgements}}
The Kallaama project was made possible thanks to the financial support from the Lacuna Fund\footnote{\url{https://lacunafund.org/}}, the world’s first collaborative effort to fund labelled data for social impact. Lacuna Fund promotes creation, expansion, and maintenance of labelled datasets in three domain areas with key needs: agriculture, health, and languages.  \\
The authors would also like to acknowledge and thank the linguists and data scientist interns  involved in the project.

\section{Bibliographical References}\label{sec:reference}

\bibliographystyle{lrec-coling2024-natbib}
% \bibliography{lrec-coling2024-example}
\bibliography{rail2024}

\begin{thebibliography}{4}
\expandafter\ifx\csname natexlab\endcsname\relax\def\natexlab#1{#1}\fi

\bibitem[{Conneau et~al.(2023)Conneau, Ma, Khanuja, Zhang, Axelrod, Dalmia,
  Riesa, Rivera, and Bapna}]{conneau2023fleurs}
Conneau, Alexis and Ma, Min and Khanuja, Simran and Zhang, Yu and Axelrod, Vera
  and Dalmia, Siddharth and Riesa, Jason and Rivera, Clara and Bapna, Ankur.
  2023.
\newblock \href {https://doi.org/10.5281/zenodo.7561858} {\emph{Fleurs:
  Few-shot learning evaluation of universal representations of speech}}.
\newblock 2022 IEEE Spoken Language Technology Workshop (SLT).
\newblock PID
  \href{https://huggingface.co/datasets/google/fleurs}{https://huggingface.co/datasets/google/fleurs}.

\bibitem[{Djiba(2021)}]{waxalGalsenAI2022}
Djiba, Daouda Tandiang. 2021.
\newblock \href {https://doi.org/10.5281/zenodo.7561858} {\emph{Keyword
  Spotting with African Languages}}.
\newblock Zenodo.
\newblock PID
  \href{https://doi.org/10.5281/zenodo.7561858}{https://doi.org/10.5281/zenodo.7561858}.

\bibitem[{Gauthier et~al.(2016)Gauthier, Besacier, Voisin, Melese, and
  Elingui}]{GAUTHIER16.460}
Gauthier, Elodie and Besacier, Laurent and Voisin, Sylvie and Melese, Michael
  and Elingui, Uriel Pascal. 2016.
\newblock \emph{Collecting Resources in Sub-Saharan African Languages for
  Automatic Speech Recognition: a Case Study of Wolof}.
\newblock European Language Resources Association (ELRA).
\newblock PID
  \href{https://hal.science/hal-01350037}{https://hal.science/hal-01350037}.

\bibitem[{Nelson(2022)}]{waxalMultilingual2022}
Nelson, Perry. 2022.
\newblock \emph{Waxal Speech Data Resources}.
\newblock PID
  \href{https://github.com/Waxal-Multilingual}{https://github.com/Waxal-Multilingual}.

\end{thebibliography}


\begin{thebibliography}{18}
\expandafter\ifx\csname natexlab\endcsname\relax\def\natexlab#1{#1}\fi

\bibitem[{Adebara and Abdul-Mageed(2022)}]{adebara2022afrocentricNlp}
Ife Adebara and Muhammad Abdul-Mageed. 2022.
\newblock \href {https://doi.org/10.18653/v1/2022.acl-long.265} {Towards
  afrocentric {NLP} for {A}frican languages: Where we are and where we can go}.
\newblock In \emph{Proceedings of the 60th Annual Meeting of the Association
  for Computational Linguistics (Volume 1: Long Papers)}, pages 3814--3841,
  Dublin, Ireland. Association for Computational Linguistics.

\bibitem[{Aker(2011)}]{aker2011dial}
Jenny~C Aker. 2011.
\newblock Dial {"A"} for agriculture: a review of information and communication
  technologies for agricultural extension in developing countries.
\newblock \emph{Agricultural economics}, 42(6):631--647.

\bibitem[{ANSD(2021)}]{ansd2021poverty}
ANSD. 2021.
\newblock \href
  {https://www.ansd.sn/sites/default/files/2022-11/Rapport-final-EHCVM-vf-Senegal.pdf}
  {{Enquête harmonisée sur les conditions de vie des ménages (EHCVM)}}.
\newblock Technical report, {Agence Nationale de la Statistique et de la
  Démographie (ANSD)}.

\bibitem[{Ciss{\'e}(2005)}]{cisse2005langues}
Mamadou Ciss{\'e}. 2005.
\newblock \href {https://au-senegal.com/IMG/pdf/doc-109pdf-33f96.pdf} {Langues,
  {\'e}tat et soci{\'e}t{\'e} au {S}{\'e}n{\'e}gal}.
\newblock \emph{SudLangues. Revue {\'e}lectronique internationale de Sciences
  du langage}, 5(1):99--133.

\bibitem[{Creissels(2019)}]{creissels2019morphology}
Denis Creissels. 2019.
\newblock \href {http://www.deniscreissels.fr/public/Creissels-morphNC.pdf}
  {{Morphology in Niger-Congo languages}}.
\newblock In \emph{Oxford Research Encyclopedia of Linguistics}.

\bibitem[{{DAPSA}(2023)}]{agri_report_2023}
{DAPSA}. 2023.
\newblock \href
  {https://www.dapsa.gouv.sn/sites/default/files/Rapport_EAA%202022_2023_v1.pdf}
  {{Rapport de l’Enquête Agricole Annuelle (EAA) 2022-2023}}.
\newblock Technical report, {Direction de l'Analyse, de la Prévision et des
  Statistiques Agricoles}.

\bibitem[{Dione(2016)}]{dione2016contacts}
Amadou Dione. 2016.
\newblock \href {https://www.theses.fr/2016GREAL023} {\emph{Contacts et
  valorisation du s{\'e}r{\`e}re et du wolof, langues nationales du
  S{\'e}n{\'e}gal: Pratiques langagi{\`e}res et usages en ligne}}.
\newblock Ph.D. thesis, Universit{\'e} Grenoble Alpes (ComUE).

\bibitem[{Joshi et~al.(2020)Joshi, Santy, Budhiraja, Bali, and
  Choudhury}]{joshi2020state}
Pratik Joshi, Sebastin Santy, Amar Budhiraja, Kalika Bali, and Monojit
  Choudhury. 2020.
\newblock \href {http://aclanthology.lst.uni-saarland.de/2020.acl-main.560.pdf}
  {The state and fate of linguistic diversity and inclusion in the {NLP}
  world}.
\newblock In \emph{Proceedings of the 58th Annual Meeting of the Association
  for Computational Linguistics}, pages 6282--6293.

\bibitem[{Leclerc(2023)}]{leclercSenegal}
Jacques Leclerc. 2023.
\newblock \href {https://www.axl.cefan.ulaval.ca/afrique/senegal.htm}
  {Sénégal}.
\newblock \emph{L'aménagement linguistique dans le monde}.

\bibitem[{Medhi et~al.(2011)Medhi, Patnaik, Brunskill, Gautama, Thies, and
  Toyama}]{medhi2011appdesign}
Indrani Medhi, Somani Patnaik, Emma Brunskill, SN~Nagasena Gautama, William
  Thies, and Kentaro Toyama. 2011.
\newblock \href {https://dl.acm.org/doi/pdf/10.1145/1959022.1959024} {Designing
  mobile interfaces for novice and low-literacy users}.
\newblock \emph{ACM Transactions on Computer-Human Interaction (TOCHI)},
  18(1):1--28.

\bibitem[{OIF(2022)}]{oif2022usage}
OIF. 2022.
\newblock \href
  {https://observatoire.francophonie.org/wp-content/uploads/2022/10/Rapport-La-langue-francaise-dans-le-monde_VF-2022.pdf}
  {La langue française dans le monde}.
\newblock Technical report, Organisation International de la Francophonie.

\bibitem[{Renaudier(2012)}]{renaudier2012derivation}
Marie Renaudier. 2012.
\newblock \href
  {http://theses.univ-lyon2.fr/documents/lyon2/2012/renaudier_m/pdfAmont/renaudier_m_these.pdf}
  {\emph{D{\'e}rivation et valence en sereer. Vari{\'e}t{\'e} de {Mar Lodj}
  ({S}{\'e}n{\'e}gal)}}.
\newblock Ph.D. thesis, Universit{\'e} Lumi{\`e}re Lyon 2.

\bibitem[{Robert(2022)}]{robert2022wolof}
St{\'e}phane Robert. 2022.
\newblock \href
  {https://corporan.huma-num.fr/Archives/WOL/PDF/WOL_SR_AWOLOF%20GRAMMATICAL%20SKETCH_PREPRINT.PDF}
  {Wolof: a grammatical sketch}.
\newblock In Friederike L{\"u}pke, editor, \emph{The Oxford guide to the
  Atlantic languages of West Africa}. Oxford University Press, Oxford.

\bibitem[{Ruder et~al.(2022)Ruder, Vuli{\'c}, and S{\o}gaard}]{ruder2022bias}
Sebastian Ruder, Ivan Vuli{\'c}, and Anders S{\o}gaard. 2022.
\newblock \href {https://doi.org/10.18653/v1/2022.findings-acl.184} {Square one
  bias in {NLP}: Towards a multi-dimensional exploration of the research
  manifold}.
\newblock In \emph{Findings of the Association for Computational Linguistics:
  ACL 2022}, pages 2340--2354, Dublin, Ireland. Association for Computational
  Linguistics.

\bibitem[{{UNESCO Institute for Statistics}(2023)}]{uis2023}
{UNESCO Institute for Statistics}. 2023.
\newblock \href {http://data.uis.unesco.org} {Education: Number of
  illiterates}.
\newblock Last consulted on 13 Feb 2024 07:49 UTC (GMT).

\bibitem[{van Esch et~al.(2022)van Esch, Lucassen, Ruder, Caswell, and
  Rivera}]{van2022writing}
Daan van Esch, Tamar Lucassen, Sebastian Ruder, Isaac Caswell, and Clara
  Rivera. 2022.
\newblock \href
  {http://lrec-conf.org/proceedings/lrec2022/pdf/2022.lrec-1.538.pdf} {Writing
  system and speaker metadata for 2,800+ language varieties}.
\newblock In \emph{Proceedings of the Thirteenth Language Resources and
  Evaluation Conference}, pages 5035--5046.

\bibitem[{{World Bank}(2021)}]{wdr2021data}
{World Bank}. 2021.
\newblock \href {https://www.worldbank.org/en/publication/wdr2021} {World
  development report 2021: Data for better lives}.

\bibitem[{Zelezny-Green et~al.(2018)Zelezny-Green, Vosloo, Conole
  et~al.}]{zelezny2018digital}
Ronda Zelezny-Green, Steven Vosloo, Gr{\'a}inne Conole, et~al. 2018.
\newblock \href {https://unesdoc.unesco.org/ark:/48223/pf0000261791} {Digital
  inclusion for low-skilled and low-literate people: a landscape review}.

\end{thebibliography}

\section{Language Resource References}
\label{lr:ref}
\bibliographystylelanguageresource{lrec-coling2024-natbib}
\bibliographylanguageresource{languageresource}

% \newpage
\appendix

%% For appendix (camera-ready paper autorized only):
\section{Details on speech transcription mistakes\label{app:mistakes_details}}

\subsection{Wolof transcriptions}
Main mistakes mentioned are: 
\begin{itemize}
    \item failure to respect certain vowel lengthenings (e.g.: word "mbool\textbf{ee}m" written instead of "mbool\textbf{e}m");
    \item failure to respect consonant gemination internally and in the final position for a number of words (e.g.: "lo\textbf{pp}aalëb" instead of "lo\textbf{p}aalëp");
    \item confusion between plosives consonants in final position of certain words (/p/ \textit{versus} /b/, /k/ \textit{versus} /g/).  
\end{itemize}

\subsection{Pulaar transcriptions}
Main mistakes reported in the Pulaar transcriptions are the following:
\begin{itemize}
    \item no distinction between the consonant \texthtd{} and the consonant \texthtb{} (e.g.: "hee\texthtd{}i" instead of "he\texthtb{}i");
    \item use of a simple consonant instead of a prenasal consonant (e.g.: "\textbf{j}iiya" instead of "\textbf{nj}iyaa");
    \item concatenation of a noun and its article (e.g.: "yim\texthtb{}e\texthtb{}e" instead of "yim\texthtb{}e \texthtb{}ee");
    \item confusion in the vowel lengthening (e.g.: "d\textbf{ee}m\textbf{o}wo" instead of "d\textbf{e}m\textbf{oo}wo"; "r\textbf{ee}m\textbf{o}\texthtb{}e\texthtb{}\textbf{e}" instead of "r\textbf{e}m\textbf{oo}\texthtb{}e \texthtb{}\textbf{ee}").
\end{itemize}

\subsection{Sereer transcriptions}
In the Sereer transcriptions, phonetical, morphological and syntactical mistakes were found. 
Notably: 
\begin{itemize}
    \item pre-nasalised consonants ("/nd/", "/mb/", "/nj/", "/ng/") used instead of glotalised or nasal consonants ("\texthtb{}", "\textipa{\!d}", "\textipa{N}");
    \item vowel lengthening not written ("ref\textbf{e}" instead of "ref\textbf{ee}","m\textbf{a}ga a mb\textbf{a}g o njirñ\textbf{a}" instead of "m\textbf{aa}ga a mb\textbf{aa}g o njirñ\textbf{aa}"; 
    \item noun and class pronoun are detached as in "xa qol axe" written "xa qola xe".
\end{itemize}

\section{Speech dataset details\label{app:speech_datasets_details}}

Rows explanation of Table \ref{tab:whole_dataset_details} and Table \ref{tab:checked_dataset_details}:
\begin{itemize}
    \item "Min (sec.)" is the minimum duration of an audio file in the given dataset.
    \item "Max (sec.)" is the maximum duration of an audio file in the given dataset.
    \item "Mean (sec.)" is the average duration of all the audio files in the given dataset.
    \item "Total audio" is the total duration of the audio set.
    \item "Total speech" is the total speech duration of the audio set.
    \item "Female speech" is the total speech duration of female speakers within the audio set. 
    \item "Male speech" is the total speech duration of male speakers within the audio set.
    \item "Female speech ratio" is the percentage of speech duration of female speakers within the audio set.
    \item "Male speech ratio" is the percentage of speech duration of male speakers within the audio set.
    \item "\#Turn-taking" is the number of speaker turn-takings in the whole audio set.
    \item "\#Files" is the total number of recordings and transcriptions in the speech dataset.
\end{itemize}

"Total speech", "Female speech", "Male speech" and "\#Turn-taking" durations have been computed from the Transcriber (.trs) files. 
This information should be treated with caution, as it depends on the accuracy of the annotations made by the transcribers.
All the others information in the table are computed from audio files (.wav).

\subsection{Whole set\label{app:whole_dataset_details}}

Table \ref{tab:whole_dataset_details} gives some statistics on the whole speech dataset. \\

\begin{table}[htbp]
    \centering
    \begin{adjustbox}{width=1\columnwidth}
    \begin{tabular}{l|r|r|r}
        \textbf{Dataset statistics} & \textbf{Wolof} & \textbf{Pulaar} & \textbf{Sereer} \\ \hline
        \textbf{Min} (sec.) & 21 & 20 & 25 \\ 
        \textbf{Max} (sec.) & 3014 & 3033 & 3461 \\ 
        \textbf{Mean} (sec.) & 1299 & 1384 & 1306 \\ 
        \textbf{Total audio} (hh:mm:ss) & 55:11:41 & 31:55:10 & 38:12:10 \\ 
        \textbf{Total speech}* (hh:mm:ss) & 51:08:50 & 30:06:43 & 36:23:37 \\ 
        \textbf{Female speech}* (hh:mm:ss) & 05:12:41 & 04:05:07 & 10:12:10 \\ 
        \textbf{Male speech}* (hh:mm:ss) & 45:56:09 & 26:01:36 & 26:11:26 \\ 
        \textbf{Female speech ratio} (\%) & 10.19 & 13.57 & 28.03 \\ 
        \textbf{Male speech ratio} (\%) & 89.81 & 86.43 & 71.97 \\ 
        \textbf{\#Turn-taking}* & 46,907 & 16,558 & 9,007 \\ 
        \textbf{\#Files} & 306 & 166 & 210 \\ 
    \end{tabular}
    \end{adjustbox}
    \raggedright \textit{\small *extracted from annotations}
    \caption{Details about Kallaama speech dataset}
    \label{tab:whole_dataset_details}
\end{table}

\subsection{Checked set\label{app:checked_dataset_details}}

Table \ref{tab:checked_dataset_details} gives some statistics on the checked subpart of the speech dataset. \\

\begin{table}[htbp]
    \centering
    \begin{adjustbox}{width=1\columnwidth}
    \begin{tabular}{l|r|r|r}
        \textbf{Dataset statistics} & \textbf{Wolof} & \textbf{Pulaar} & \textbf{Sereer} \\ \hline
        \textbf{Min} (sec.) & 21 & 117 & 444 \\ 
        \textbf{Max} (sec.) & 2849 & 3033 & 2907 \\ 
        \textbf{Mean} (sec.) & 1283 & 1472 & 1250 \\ 
        \textbf{Total audio} (hh:mm:ss) & 12:49:35 & 11:02:28 & 11:06:52 \\ 
        \textbf{Total speech}* (hh:mm:ss) & 11:47:34 & 10:56:15 & 10:51:33 \\ 
        \textbf{Female speech}* (hh:mm:ss) & 01:27:00 & 01:08:09 & 03:12:29 \\ 
        \textbf{Male speech}* (hh:mm:ss) & 10:20:33 & 09:48:06 & 07:39:03 \\ 
        \textbf{Female speech ratio} (\%) & 12.30 & 10.39 & 29.54 \\ 
        \textbf{Male speech ratio} (\%) & 87.70 & 89.61 & 70.46 \\ 
        \textbf{\#Turn-taking}* & 11,968 & 3,583 & 1,796 \\ 
        \textbf{\#Files} & 72 & 54 & 64 \\ 
    \end{tabular}
    \end{adjustbox}
    \raggedright \textit{\small *extracted from annotations}
    \caption{Details about checked part of Kallaama speech dataset}
    \label{tab:checked_dataset_details}
\end{table}

\newpage
\section{Recordings types\label{app:program_types}}

The recordings are from various types of programmes and are rated on a scale of 1 to 5 based on their potential complexity for speech processing. \\
This rating is subjective and takes into account factors such as recording duration, number of talking speakers, and recording conditions. \\
A rating of 1 indicates relatively low complexity, while a rating of 5 indicates relatively high complexity. This ID is the first number composing the name of the files. \\
Table \ref{tab:program_types} shows the number of recordings per programme type, for each language set.

\begin{table}[hb]
    \centering
    \begin{adjustbox}{width=1\columnwidth}
    \begin{tabular}{c|l|r|r|r}
        \textbf{Type ID} & \textbf{Type} & \textbf{Wolof} & \textbf{Pulaar} & \textbf{Sereer} \\ \hline
        1 & push message & 9 & 1 & 0 \\ 
        2 & voice message & 0 & 0 & 14 \\ 
        3 & interview & 22 & 10 & 15 \\ 
        4 & radio show & 120 & 72 & 67 \\ 
        5 & focus group & 2 & 0 & 9 \\ 
    \end{tabular}
    \end{adjustbox}
    \caption{Number of recordings per programme type, for each language dataset}
    \label{tab:program_types}
\end{table}

\end{document}